\documentclass[10pt]{article}
\usepackage{spconf,amsmath,epsfig}

\usepackage{lettrine}
\usepackage[numbers]{natbib}
\usepackage[usenames,dvipsnames]{color}

\usepackage[acronym, shortcuts]{glossaries}
\newacronym{ml}{ML}{Machine Learning}
\newacronym{ai}{AI}{Artificial Intelligence}
\newacronym{dl}{DL}{Deep Learning}
\newacronym{xai}{XAI}{eXplainable AI}

\newacronym{eo}{EO}{Earth Observation}
\newacronym{rs}{RS}{Remote Sensing}
\newacronym{s2}{S2}{Sentinel-2}

\newacronym[longplural={vegetation indices}]{vi}{VI}{vegetation index}
\newacronym{ndvi}{NDVI}{normalized difference vegetation index}
\newacronym{nndvi}{n-NDVI}{narrow normalized difference vegetation index}
\newacronym{ndmi}{NDMI}{normalized difference water index}
\newacronym{ndre}{NDRE}{normalized difference red edge}
\newacronym{ireci}{IRECI}{Inverted Red-Edge Chlorophyll Index}

\newacronym{ndwi}{NDWI}{normalized difference water index}
\newacronym{cig}{CIG}{chlorophyll index green}
\newacronym{evi}{EVI}{enhanced vegetation index}
\newacronym{savi}{SAVI}{soil-adjusted vegetation index}
\newacronym{dsi}{DSI}{drought stress index}
\newacronym{ttvi}{TTVI}{Transformed Triangular Vegetation Index}
\newacronym{nmdi}{NMDI}{Normalized Difference Moisture Index}

\newacronym{nir}{NIR}{near-infrared}
\newacronym{nnir}{n-NIR}{narrow near-infrared}
\newacronym{swir}{SWIR}{short-wave infrared}
\newacronym{re}{RE}{red-edge}

\newacronym{rf}{RF}{random forest}
\newacronym{gpr}{GPR}{gaussian process regression}
\newacronym{svm}{SVM}{support vector machine}
\newacronym{mlp}{MLP}{multilayer perceptron}
\newacronym{dnn}{DNN}{deep neural network}
\newacronym{rnn}{RNN}{recurrent neural network}
\newacronym{cnn}{CNN}{convolutional neural network}
\newacronym{dfnn}{DFNN}{deep forward neural network}
\newacronym{lstm}{LSTM}{Long short-term memory neural network}
\newacronym{gru}{GRU}{Gated Recurrent Unit}
\newacronym{gbdt}{GBDT}{gradient-boosted decision tree}
\newacronym{pca}{PCA}{Principal Component Analysis}

\newacronym{oa}{OA}{overall accuracy}
\newacronym{rmse}{RMSE}{root mean square error}
\newacronym{r2}{$\text{R}^2$}{coefficient of determination}
\newacronym{pp}{$pp$}{percentage points}

\newacronym{svs}{SVS}{Shapley Value Sampling}

\setacronymstyle{long-short}

\title{XAI-GUIDED ENHANCEMENT OF VEGETATION INDICES\\ FOR CROP MAPPING}

\twoauthors
  {Hiba Najjar, Francisco Mena\sthanks{H.Najjar and F.Mena acknowledge support through a scholarship from
the University of Kaiserslautern-Landau.\\ This work is accepted as a conference paper at IGARSS2024.}}
  {University of Kaiserslautern-Landau \\ Kaiserslautern\\ Germany}
  {Marlon Nuske, Andreas Dengel}
  {German Research Center for Artificial Intelligence \\ Kaiserslautern\\ Germany}

\begin{document}
%
\maketitle

\begin{abstract}
    Vegetation indices allow to efficiently monitor vegetation growth and agricultural activities.  Previous generations of satellites were capturing a limited number of spectral bands, and a few expert-designed vegetation indices were sufficient to harness their potential. New generations of multi- and hyperspectral satellites can however capture additional bands, but are not yet efficiently exploited. 
    In this work, we propose an explainable-AI-based method to select and design suitable vegetation indices. We first train a deep neural network using multispectral satellite data, then extract feature importance to identify the most influential bands. We subsequently select suitable existing vegetation indices or modify them to incorporate the identified bands and retrain our model.
    We validate our approach on a crop classification task. Our results indicate that models trained on individual indices achieve comparable results to the baseline model trained on all bands, while the combination of two indices surpasses the baseline in certain cases.

\end{abstract}

\section{Introduction}\label{sec:intro}

    In recent years, an increasing number of studies have employed \gls{ml} and \gls{dl} techniques to harness remote sensing data for multiple applications related to the sustainable development goals \cite{ferreira2020monitoring}. 
    While such models are proficient in processing raw satellite bands, a common data engineering practice in this field involves the utilization of \glspl{vi}. 
    Ratios, differences, and derivatives between reflectance values from different spectral wavelengths can enhance the spectral signals associated with vegetation characteristics of interest, given that the original measurements of spectral reflectance constitute a mixed signal comprising vegetation canopies, shadows, soils, and other components present on the land surface \cite{zeng2022optical}.
    While some \glspl{vi}
    are commonly used for crop monitoring, the selection of the most suitable vegetation index is not always straightforward \cite{zeng2022optical}. Instead, the initial step involves identifying the sensitive wavelengths and corresponding \glspl{vi} for their optimal utilization.
    
    An advantage of using \gls{dl} lies in the model's inherent capability to automatically extract crop-related features and discern interactions between raw bands. 
    To extract scientific insights encoded in the model, \gls{xai} techniques can uncover the inner workings of the model, facilitating an understanding of how individual satellite bands contribute to its predictions \cite{ras2022explainable}. 
    Regarding the \gls{s2} multispectral instruments in particular, they stand out as one of the few remote sensors with the capacity to capture \gls{re} wavelengths between 700 and 800nm. Notably, the additional \gls{re} bands remain under-explored for their potential to enhance crop classification through vegetation indices \cite{misra2020status}. Furthermore, \gls{swir} bands, typically used for water monitoring, have also received little attention in exploring their efficacy to track vegetation cover and its phenology \cite{misra2020status}.
    In this paper, we introduce an approach that leverages explainability methods to identify relevant bands and improve the use of \glspl{vi}. We validate our approach on a crop classification task. (The code will be made publicly available upon paper acceptance.)

\section{Methodology}\label{sec:method}

    \subsection{Crop Dataset}

        In Sub-Suharan Africa, extreme food insecurity and malnutrition are prevalent in multiple countries. 
        In this study, we leverage \gls{s2} data from Ghana and South Sudan to address this task. 
        The corresponding public datasets used contains satellite image time series captured between January and December 2016 at a 10m resolution, and are labeled with multiple land cover classes \cite{m2019semantic}. For our study, we merge the two datasets and retain only the pixels corresponding to crops. We focus our work on classes with more than 10,000 labeled pixels: sorghum, maize, rice, groundnut, soybean, and yam. 
        Table \ref{tab:data} presents the data distribution in each country. 
        We partition 5\% of the data for validation, ensuring that pixels originating from the same satellite image patch are exclusively utilized for either training or validation but not both.

        \begin{table}[h]
        \centering
        \footnotesize
        \caption{Pixel count per crop type.}
            \begin{tabular}{|c|c|c|c|}
            \hline
            Crop & Total & Ghana & S-Sudan \\ \hline
            Maize & 329,847 & 322,767 & 7,080 \\ 
            Groundnut & 101,314 & 96,371 & 4,943 \\ 
            Rice & 98,986 & 93,908 & 5,078 \\ 
            Soybean & 67,638 & 67,638 & - \\ 
            Sorghum & 65,185 & 8,352 & 56,833 \\
            Yam & 22,091 & 22,091 & - \\ \hline
            \end{tabular}
        \label{tab:data}
        \end{table}

    \subsection{Exploiting spectral attributions}\label{ssec:svs}

        Feature attribution methods are explanation techniques that provide interpretations for individual predictions. These methods assign sensitivity or contribution scores to each input feature, quantifying their relative importance to the model's prediction \cite{shap_lundberg2017unified}. 
        In our experiments, we use the \gls{svs} to estimate feature attributions \cite{svs_strumbelj2010efficient}.  
        \gls{svs} is grounded in cooperative game theory, which provides a solid theoretical foundation, unlike many other methods \cite{shap_lundberg2017unified}. 
        Its robustness has being quantitatively evaluated in the context of a regression task based on time series of satellite data, and has shown superior stability against several other techniques \cite{najjarFeatureAttribution2023}.

        The results of the spectral attribution are used to improve the selection of \glspl{vi} for the crop mapping task. 
        We first interpret the model trained on the ten satellite bands by estimating the attributions for a maximum of 10,000 correctly classified pixels from each crop. 
        The features are grouped over the spectral dimension to compute a single attribution value for the time series of each band. 
        The negative attributions are suppressed to only consider positive contributions to a given class \cite{selvaraju2017grad}. 
        To standardize the results, we scale the attributions so that the summation of attributions per pixel equals 1, 
        before averaging them both globally and per class.
        Subsequently, we use these importance values to select \glspl{vi} that account for these bands, and adjust existing indices as needed. 
        The model is then retrained by replacing the satellite bands with individual indices or binary combinations.

    \subsection{Experimental setup}
    
        We use ten bands from \gls{s2} data for our analysis, namely the blue (B02), green (B03), red (B04), three \gls{re} bands (B05, B06, B07), \gls{nir} (B08), \gls{nnir} (B8A), and two \gls{swir} (B11, B12) bands. 
        An additional channel, indicating the cloud coverage of the image, is stacked to these bands and used in all our experiments.
        
        Regarding the modeling, we rely on recurrent neural networks, which have successfully been used to analyze temporal satellite data 
        \cite{jia2017incremental, sharma2018land, garnot2019time, mou2018learning}.
        We opt for the \gls{gru}, introduced in \cite{chung2014empirical}, due to its moderate number of parameters and its proven effectiveness in remote sensing applications \cite{garnot2019time, interdonato2019duplo, mou2017deep}. 
        The time series of each pixel are pre-padded to a fixed sequence length of 228, to account for the longest time series in the dataset, before being supplied to the model.
        To handle the unbalanced labels in the data, we use a weighted sampler during training. 
        We evaluate all models using the \gls{oa} and F1 scores. The \gls{oa} is the percentage of correctly classified pixels across all classes, while the F1 score is the weighted average of class specific F1 scores. We also report the accuracies per class.

\section{Results}\label{sec:results}

    \subsection{Spectral attributions}
        We train the \gls{gru}-based model using the satellite bands and present the evaluation results on the validation set in the second column of Table \ref{tab:cm_res}. 
        This baseline model achieved a score of 67\% on both the \gls{oa} and F1 metrics. In individual classes, high accuracies of 84\% and 86\% were attained for rice and sorghum, respectively, while yam exhibited the lowest score at 27\%. This could be attributed to the relatively small number of pixels in this class. Notably, the largest two classes did not necessarily exhibit the best performance, suggesting that the performance gaps are not solely due to the size of each class.

        \begin{table*}[]
    \centering
    \footnotesize
    \caption{Experimental results of all trained models. The best score in each experimental group is in bold.}
        \begin{tabular}{|l|c|ccccccc|ccccccc|}
            \hline
            \multicolumn{1}{|c|}{} & \multicolumn{1}{c|}{S2} & \multicolumn{7}{c|}{Single VI} & \multicolumn{7}{c|}{Two VIs} \\ \hline
            S2 & x & — & — & — & — & — & — & — & — & — & — & — & — & — & —  \\
            NDVI & — & x & — & — & — & — & — & — & x & x & x & — & — & — & —  \\
            nNDVI & — & — & x & — & — & — & — & — & — & — & — & — & — & x & x  \\
            NDRE & — & — & — & x & — & — & — & — & x & — & — & x & x & x & —  \\
            NDRE2 & — & — & — & — & x & — & — & — & — & x & — & — & — & — & —  \\
            NDRE3 & — & — & — & — & — & x & — & — & — & — & x & — & — & — & —  \\
            NDMI & — & — & — & — & — & — & x & — & — & — & — & x & — & — & x  \\
            NDMI2 & — & — & — & — & — & — & — & x & — & — & — & — & x & — & —  \\
            \hline
            OA & 0.67 & 0.62 & 0.62 & 0.61 & 0.56 & 0.51 & \textbf{0.65} & 0.63 & 0.64 & 0.62 & 0.62 & 0.67 & \textbf{0.70} & 0.61 & 0.68 \\
            F1 & 0.67 & 0.63 & 0.62 & 0.61 & 0.57 & 0.52 & \textbf{0.65} & 0.64 & 0.65 & 0.62 & 0.63 & 0.67 & \textbf{0.70} & 0.62 & 0.69 \\
            \hline
            Maize & 0.65 & \textbf{0.66} & 0.61 & 0.65 & 0.54 & 0.41 & 0.62 & 0.60 & 0.67 & 0.60 & 0.61 & 0.63 & \textbf{0.70} & 0.61 & 0.66 \\
            Groundnut & 0.51 & 0.45 & 0.51 & 0.48 & 0.45 & 0.44 & \textbf{0.57} & 0.50 & 0.49 & 0.51 & 0.57 & \textbf{0.69} & 0.61 & 0.44 & 0.61 \\
            Rice & 0.84 & 0.64 & 0.70 & 0.62 & 0.62 & 0.64 & 0.73 & \textbf{0.76} & 0.66 & 0.74 & 0.66 & 0.77 & \textbf{0.83} & 0.71 & 0.81 \\
            Soybean & 0.48 & 0.49 & 0.42 & 0.34 & 0.38 & 0.49 & 0.41 & \textbf{0.50} & \textbf{0.50} & 0.37 & 0.42 & 0.35 & 0.38 & 0.43 & 0.49 \\
            Sorghum & 0.86 & 0.84 & 0.84 & 0.81 & 0.84 & 0.83 & \textbf{0.87} & 0.86 & 0.85 & 0.83 & 0.82 & 0.85 & \textbf{0.88} & 0.82 & 0.87 \\
            Yam & 0.27 & 0.23 & 0.21 & 0.23 & 0.29 & 0.34 & \textbf{0.38} & 0.27 & 0.22 & 0.30 & 0.23 & 0.31 & 0.32 & \textbf{0.33} & 0.23  \\ 
            \hline
        \end{tabular}
    \label{tab:cm_res}
\end{table*}
        
        \begin{figure}[t]
            \centering
            \includegraphics[width=1\columnwidth]{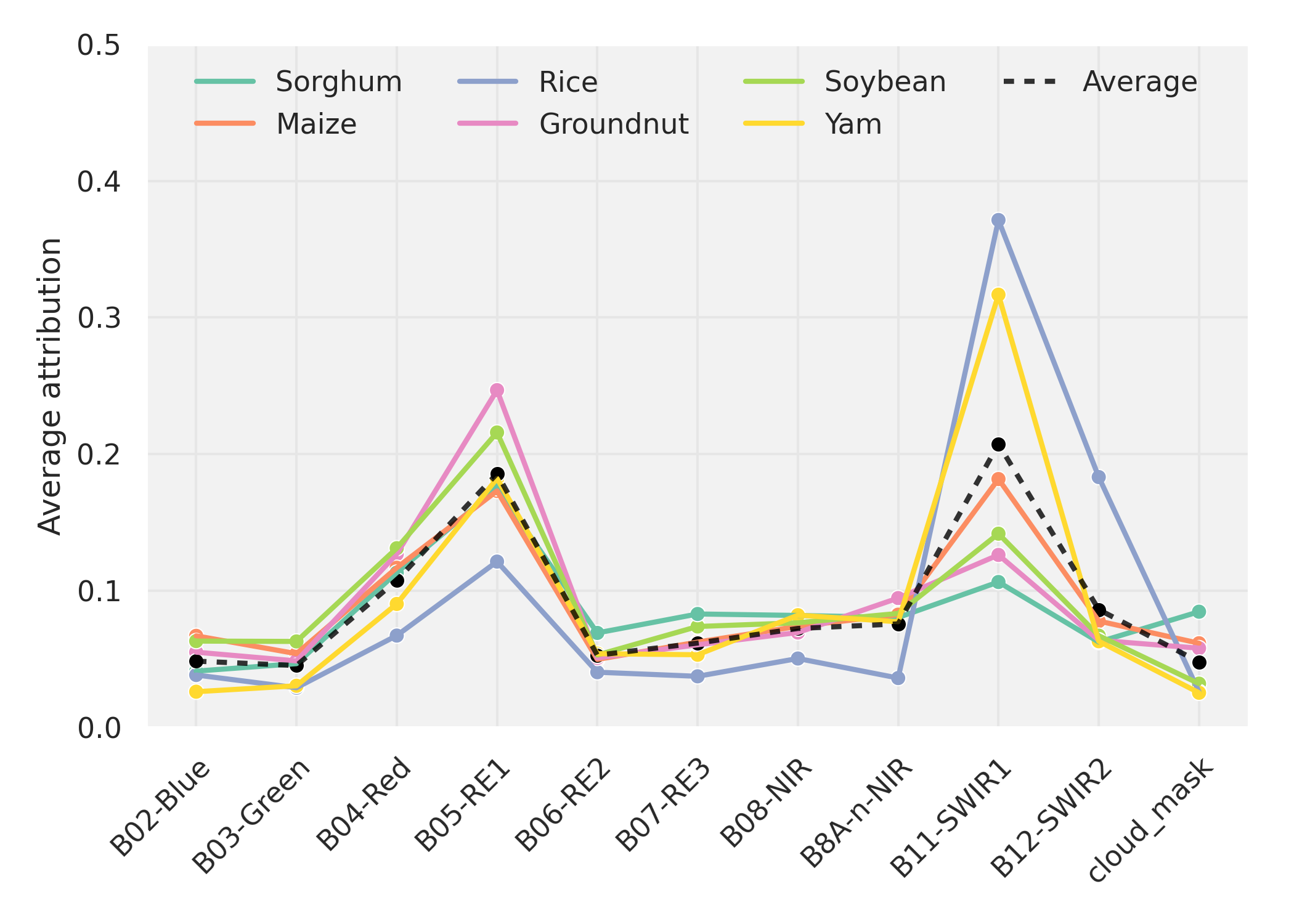}
            \caption{Global and crop-specific spectral attributions of the model trained on the ten satellite bands.}
            \label{fig:spec_attr}
        \end{figure}

        We interpret the baseline model following the procedure described in \ref{ssec:svs}, and visualize the corresponding results in Figure \ref{fig:spec_attr}.
        Starting with the global average attribution line,  \gls{swir}1 and \gls{re}1 rank at the top with around 20\% of the total importance, followed by the red, \gls{swir}2, \gls{nnir}, and \gls{nir} bands, in the descendant order of their respective importance. The remaining bands exhibit a less significant importance.
        Notably, the relatively small importance of the cloud mask across all classes indicates that the model is not biased by this channel for the identification of any specific crop.

        To analyze the results crop-wise, 
            groundnut and soybean highly rely on the first \gls{re} band, followed by the red and \gls{swir}1 bands. Sorghum has a similar attribution pattern. Rice has an additional particular dependence on the \gls{swir}2 band. 
            Rice and yam identification significantly rely on the first \gls{swir} band, followed by \gls{re}1. All the remaining bands have each less than 10\% of the total importance.
            Maize crop classification is sensitive to the first \gls{swir} and \gls{re} bands, followed by the red band.

        These results highlight the relevance of \gls{re}1 and \gls{swir}1 bands for crop mapping and complement the findings of earlier studies.
        Yi et al. \cite{yi2020crop} assessed the importance of \gls{s2} bands on the same task and found that \gls{re}1 and \gls{swir}1 bands are more efficient in identifying crops than other bands in the Shiyang River Basin in China.
        Similarly, Liu et al. \cite{liu2021comprehensive} found that \gls{re} and \gls{swir} bands of \gls{s2} had irreplaceable effects on land cover classification.

    \subsection{Enhanced usage of \glspl{vi}}

        In light of the insights gained from the importance of the satellite bands for crop mapping, we proceed with a guided selection of individual and binary combinations of \glspl{vi}.

        \begin{table}[t]
        \footnotesize
        \centering
        \caption{\glspl{vi} used for crop mapping. R, N, nN, S1, and S2 are the red, \gls{nir}, \gls{nnir}, \gls{swir}1, and \gls{swir}2, respectively.}
            \begin{tabular}{|c|c|c|}
            \hline
            VI & Formula & Reference \\ \hline
            NDVI & (N - R)/(N + R) & Rouse et al. \cite{NDVI_rouse1974monitoring}  \\
            n-NDVI & (nN - R)/(nN + R) & This paper \\
            NDRE & (N - RE1) / (N + RE1) & Gitelson \& Merzlyak\cite{NDRE_gitelson1994quantitative}\\
            NDRE2 & (N - RE2) / (N + RE2) & This paper\\
            NDRE3 & (N - RE3) / (N + RE3) & This paper\\
            NDMI & (N - S1)/(N + S1) & Wilson \& Sader \cite{NDMI_wilson2002detection}\\
            NDMI2 & (N - S2)/(N + S2) & This paper\\
            \hline
            \end{tabular}
        \label{tab:vi}
        \end{table}

        Given the significance of \gls{re}1, we include the \gls{ndre} \cite{NDRE_gitelson1994quantitative} index that uses the \gls{nir} and \gls{re}1 bands. We derive two modified indices, \gls{ndre}2 and \gls{ndre}3, by replacing the first \gls{re} channel with the second and third, respectively, to verify whether the relative performance of the three indices align with the attribution of their respective bands.
        We also incorporate the \gls{ndmi}, which uses the first \gls{swir} band, and create a modified version, \gls{ndmi}2, which uses the second \gls{swir} band, influential on rice identification.
        Additionally, we include the widely used \gls{ndvi}, and recognizing the comparable importance of \gls{nnir}, we introduce a modified index, \gls{nndvi}, where the \gls{nir} band is replaced with \gls{nnir}.

        It is important to note that only the red, green, blue, and \gls{nir} bands have a resolution of 10m, while the remaining bands were originally captured either at a 20 or 60m resolution. Therefore, we ensured that all our proposed indices contain at least one of the high-resolution bands. The formula of each index is listed in Table \ref{tab:vi}. We retrain our model using individual indices or combinations of two indices as inputs. The results are reported in Table \ref{tab:cm_res}.

        Among the models trained on a single \gls{vi}, the top-performing model is based on \gls{ndmi}, achieving an \gls{oa} score of 67\%. This model outperformed the baseline in identifying three crops: sorghum, groundnut, and yam. 
        The second-best model is based on the modified version of the same index, \gls{ndmi}2, which achieved the same class accuracy as the baseline in sorghum and yam, and performed better in soybean.
        The third-best model, based on the \gls{ndvi}, slightly outperformed the baseline on maize and soybean crops. The \gls{nndvi}
        The \gls{ndre}3-based model achieved the lowest \gls{oa} score, mainly due to its low accuracy in maize and rice crops.
        
        Among the models trained on two \glspl{vi}, the combination of \gls{ndre} and \gls{ndmi}2 achieved the highest accuracies for sorghum, maize, and rice, and outperformed the baseline in groundnut and yam crops. This combination also scored an \gls{oa} score of 70\%, 3 \gls{pp} higher than the baseline model. 
        The combination of \gls{ndmi} and \gls{nndvi} also demonstrated comparable performance.
        In contrast, combining \gls{ndre} and \gls{nndvi} had the worst overall performance, mainly due to its low accuracy in rice crops, despite its higher capacity to identify yam compared to the other models.
        The combinations of \gls{ndvi}+\gls{ndre}2 and \gls{ndvi}+\gls{ndre}3 also displayed comparatively low overall performance.

\section{Discussion}\label{sec:discussion}
    
    Our overall approach of exchanging the raw satellite bands with few \glspl{vi} exhibits promising results. The best model based on a single index exhibited an \gls{oa} 2\gls{pp} lower than the baseline model, while using two indices achieved a 3\gls{pp} higher accuracy in the best case. These results highlight the potential of relying solely on one or two \glspl{vi} for crop identification, especially when carefully selected. In general, larger datasets benefit from increased input features, as they enable automatic learning of high-level features by the model. However, in medium-sized training datasets like ours, performance can be enhanced through careful input feature selection.

    As shown in Figure \ref{fig:spec_attr}, \gls{swir}1 appears to be significantly important to identify rice and yam crops, accordingly the \gls{ndmi}-based models achieves the best accuracy for yam and the second-best score for rice, among the single-index based models. Combining \gls{ndmi} with a second index also achieved high accuracies for both crops.
    We further observe that the proposed \gls{ndmi}2 achieved the best accuracies on rice compared to the other single-\gls{vi} based models. Additionally, it demonstrated the highest accuracies on sorghum, maize, and rice when combined with \gls{ndre}, outperforming all \gls{vi}-based models. 
    On the other hand, the proposed \gls{ndre}2 and \gls{ndre}3 indices performed poorly on the \gls{oa} both when used individually and when combined with \gls{ndvi}, in contrast to \gls{ndre}, which achieved high scores, particularly when combined with \gls{ndmi} or \gls{ndmi}2. 
    This observation aligns with the relative average importance of the three \gls{re} bands, as shown in Figure \ref{fig:spec_attr}, suggesting that the first band is more suitable for crop identification. 
    Nonetheless, the second and third \gls{re} bands were of higher importance for soybean and sorghum compared to the remaining crops, which is consistent with the improvement in crop-specific accuracies achieved by the \gls{ndre}2 and \gls{ndre}3-based models compared to \gls{ndre}. In contrast, when combined with \gls{ndvi}, the \gls{ndre} performs better in both crops.

    While the performance of the \gls{vi}-based modeling aligns with the attribution results conducted on the baseline model, there were some behaviors that were not easily interpretable.
        For instance, soybean identification relies significantly on the first \gls{re} band, and while \gls{re}1 and \gls{re}3 have marginal importance, according to the attribution results. Nonetheless, the soybean classification accuracy is the much worse when the model is trained with \gls{ndre}, compared to the \gls{ndre}2 and \gls{ndre}3 models.  
        Similarly, the \gls{re}1 exhibits higher importance for identifying yam crop compared to the other two bands, while the performance of the three corresponding single-\gls{vi} based models had the opposite behavior. 
    
    Overall, one limitation of our \gls{xai}-based approach is the reliability of the model. Meaningful explanation results and relevant scientific insights are conditioned by the scientific accuracy of what the model has learned during the training. Since our baseline had an \gls{oa} score of 67\%, we believe that further improvements in the model's performance can enhance its robustness, and consequently, the reliability of its attribution results.
    
    In future work, in addition to improving the performance of the baseline model, we aim to extend the dataset to cover other regions from multiple years, and validate our approach on a broader range of crop types.

\section{Conclusion}\label{sec:conslusion}

    In this paper, we identified \glspl{vi} relevant to identify each crop type, guided by spectral importance results explaining the baseline model. Our findings contribute to the growing body of evidence suggesting that the information contained within the \gls{re} and \gls{swir} bands from \gls{s2} is critical for discriminating crop types \cite{misra2020status}. Based on the explanation results, we trained several models on individual and combinations of two \glspl{vi} and demonstrated their ability to outperform the model trained on all the bands. Importantly, the performance of these models aligned with the spectral importance in crop accuracies in most cases.
    Overall, our results further indicate that combining two \glspl{vi} perform better than using a single index, and while some combinations improved the \gls{oa} over the validation set, an examination of individual crop performance reveals that an index can be highly efficient in identifying certain crops but might struggle with others.

\bibliographystyle{IEEEbib}
\bibliography{refs_initials}

\end{document}